\documentclass[journal]{IEEEtran}
\usepackage{cite}
\newcommand{\eat}[1]{}
\newcommand{\ie}{{\it{i.e.,~}}}
\newcommand{\eg}{{\it{e.g.,~}}}

\usepackage{colortbl}
\definecolor{mygray}{gray}{.9}

\newcommand{\ours}{MMGPL}
\usepackage{framed,multirow}
\usepackage{natbib}
\usepackage{setspace}
\usepackage{amssymb}
\usepackage{latexsym}
\usepackage{amsmath,amsfonts}
\usepackage{subfigure}
\usepackage{url}
\usepackage{xcolor}

\usepackage[colorlinks=true, linkcolor=red,  anchorcolor=red, citecolor=blue!30!cyan]{hyperref}
\usepackage{colortbl}

\usepackage{rotating}
\usepackage{booktabs, makecell}
\definecolor{mygray}{gray}{.9}
\definecolor{newcolor}{rgb}{.8,.349,.1}

\def\BibTeX{{\rm B\kern-.05em{\sc i\kern-.025em b}\kern-.08em
    T\kern-.1667em\lower.7ex\hbox{E}\kern-.125emX}}

\begin{document}

\title{MMGPL: Multimodal Medical Data Analysis with Graph Prompt Learning}

\author{Liang Peng, Songyue Cai, Zongqian Wu, Huifang Shang, Xiaofeng Zhu, and Xiaoxiao Li
\thanks{Corresponding author (seanzhuxf@gmail.com)}
}
\maketitle

\begin{abstract}
Prompt learning has demonstrated impressive efficacy in the fine-tuning of multimodal large models to a
wide range of downstream tasks. Nonetheless, applying existing prompt learning methods for the diagnosis
of neurological disorder still suffers from two issues: (i) existing methods typically treat all patches equally,
despite the fact that only a small number of patches in neuroimaging are relevant to the disease, and (ii) they
ignore the structural information inherent in the brain connection network which is crucial for understanding
and diagnosing neurological disorders. To tackle these issues, we introduce a novel prompt learning model
by learning graph prompts during the fine-tuning process of multimodal models for diagnosing neurological
disorders. Specifically, we first leverage GPT-4 to obtain relevant disease concepts and compute semantic
similarity between these concepts and all patches. Secondly, we reduce the weight of irrelevant patches
according to the semantic similarity between each patch and disease-related concepts. Moreover, we construct
a graph among tokens based on these concepts and employ a graph convolutional network layer to extract the
structural information of the graph, which is used to prompt the pre-trained multimodal models for diagnosing
neurological disorders. Extensive experiments demonstrate that our method achieves superior performance for
neurological disorder diagnosis compared with state-of-the-art methods and validated by clinicians.

\end{abstract}

\section{Introduction}
\label{sec1}
Neurological disorders, including Autism Spectrum Disorder (ASD) \citep{lord2018autism} and Alzheimer's Disease (AD) \citep{scheltens2021alzheimer}, severely impair subjects' social, linguistic, and cognitive abilities, and have already become serious public health issues worldwide \citep{feigin2020global}. Unfortunately, there are no definitive cures for most neurological disorders (\eg ASD and AD), so diagnosis of neurological disorder is urgently needed to promote early intervention and delay its deterioration \citep{wingo2021integrating, zhu2022interpretable}. 
Over the last decade, researchers \citep{wen2020convolutional,li2021braingnn,dvornek2019jointly} have applied various machine learning methods, such as Convolutional Neural Networks (CNN) \cite{lecun1995convolutional}, Graph Neural Networks (GNN) \cite{kipf2017semi}, and Recurrent Neural Networks (RNN) \cite{schuster1997bidirectional}, to diagnose neurological disorders.
Despite the remarkable progress made by these methods, the robustness and effectiveness of these deep learning models are difficult to ensure due to the fact that these methods are directly trained on small-scale and complicated medical datasets \citep{dinsdale2022challenges}.

Recently, multimodal large models \citep{LLaVA, PaLM, tu2023towards, wu2023towards} with extensive parameters, trained on vast datasets and diverse tasks, have exhibited remarkable generality and adaptability. As a result, multimodal large models have become a recent focal point within the field of medical data analysis. Researchers across various domains have developed disparate products such as the large language models (\eg GPT \citep{openai2023gpt}) and the large vision models (\eg SAM \citep{kirillov2023segment}). They can accelerate the development of accurate and robust models, reducing reliance on extensive labeled data \citep{zhang2023challenges}. Owing to their generality, multimodal large models hold immense potential in addressing a wide array of diagnostic tasks for neurological disorders.

However, applying these multimodal large models in the field of neurological disorders diagnosis faces significant challenges due to the diverse modalities in multimodal medical data (\eg PET and MRI), which differ greatly from natural images. To fill the gap between the pre-training tasks and downstream tasks, researchers utilize techniques such as full fine-tuning and prompt learning on pre-trained multimodal large models to address specific downstream tasks in medicine domains. Specifically, full fine-tuning is a commonly used method in transfer learning, which updates the weights of pre-trained models on task-specific data. However, full fine-tuning methods \citep{howard-ruder-2018-universal} are becoming increasingly difficult to operate when dealing with large models with massive parameters because they require fine-tuning the entire model directly. Recently, prompt learning \citep{brown2020language} has become a new fine-tuning paradigm of modern multimodal large models, which explicit instructions or adjusts only a small portion of parameters of prompts to guide the foundation model. For example, \cite{lester2021power} demonstrates that even for models of the 10 billion parameter scale, optimizing the prompt alone while keeping the model parameters fixed can achieve performance comparable to that of full fine-tuning.

Despite the impressive results demonstrated by prompt learning, it is highly task-specific, requiring researchers from different domains to design customized prompts to maximize its potential.
In the field of Natural Language Processing (NLP), previous works usually employed cloze prompts, which can be designed manually or automatically. For example, \cite{brown2020language} manually defines specific prompt templates to boost various NLP tasks. However, manually adjusting the prompt templates lacks flexibility and efficiency. To yield more flexible and efficient prompts, many researchers \citep{lu2022prompt, gao2020making, schick2020few}  proposed to discover prompts through automatic learning under supervised data.
Inspired by the successful application of prompt learning in NLP, researchers have started exploring prompt learning in computer vision (CV). For instance, VPT \citep{jia2022visual} proposed a single-modality prompt learning method that designs individual prompts for images. However, single-modality prompt learning method only prompts one modality and ignores the prompt from the other modality.
To yield more effective prompt learning, \citep{khattak2023maple, zhou2022conditional} designed to prompt both image modality and text modality.
Despite the widespread application of prompt learning methods in the fields of NLP and CV, we still face challenges when applying prompt learning to fill the gap between the pre-training multimodal large models and the diagnosis of neurological disorders.

\begin{figure*}[!t]
\centering
\includegraphics[scale=0.58]{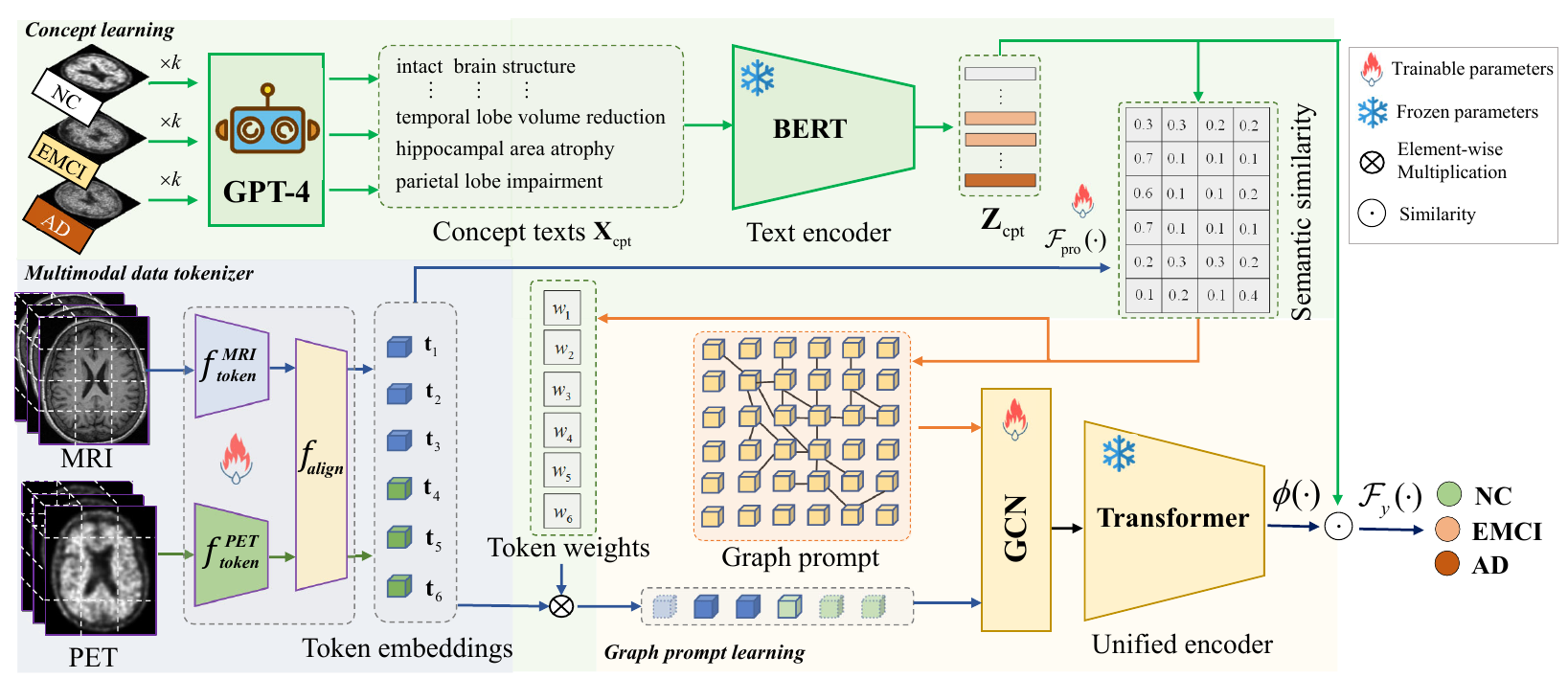}
\caption{The flowchart of the proposed \ours{} consists of three modules \ie multimodal data tokenizer (light blue block), concept learning (light green block), and graph prompt learning (light yellow block). First, \ours{} divides the multimodal medical data into multiple patches and project them into a shared embedding space (Sec. \ref{sec_tokenizer}). Second, \ours{} prompts the GPT-4 to generate disease-related concepts and further learn the weights of tokens based on the semantic similarity between tokens and concepts (Sec. \ref{sec_Concept}). Third,  \ours{} learns a graph among tokens and extracts structural information to prompt the unified encoder (Sec. \ref{sec_Graph}).  Finally, \ours{} obtains the output from the unified encoder and uses it to predict the label of the subject.
}
\label{fig_framework}
\end{figure*}

Firstly, only a small fraction of patches in neural data are related to the disease.
Unlike natural image data, the pathological regions or the regions of interest (\ie ROIs) in neural data typically only occupy a small portion of the entire data. For example, previous studies \citep{padurariu2012hippocampal, wang2006changes} have shown a significant correlation between the hippocampal area and Alzheimer's disease.
Recently, in the area of medical image segmentation, methods \citep{huang2023segment, zhang2023input} utilize region-based prompts, such as bounding boxes and clicks, to guide the fundamental model focus on ROIs within the medical images. 
Despite the promising results obtained by region-based prompts, the aforementioned operations (\eg bounding box and clicks) often rely on human interaction or object detection models which reduces the flexibility of these methods. Moreover, implementing the aforementioned operations becomes even more challenging when dealing with multimodality neural data, which may typically contains 3D tensor MRI data, time-series data, and functional connectivity (FC) networks data. Hence, it is challenging to discover useful patches when dealing with multimodality neural data.

Secondly, the structural information among patches plays a crucial role in the analysis of neurological disorders. In the field of neuroscience, researchers \citep{bullmore2009complex, fornito2015connectomics} have indicated that the brain is a complex and interconnected network/graph, and the connection topology (\ie structural information) of the brain fundamentally shapes the progression of neurological disorders. Although the transformer architecture includes a graph block (\ie composed of keys, queries, and values) for extracting global patterns, its dense nature \citep{jaszczur2021sparse} makes the model difficult to capture the structural information among patches. Additionally, the model parameters are fixed in prompt-based fine-tuning, which means that the structural information learned through key-query weights may not suitably represent the structural information relevant to neurological disorders. Thus, it is essential to extract structural information among patches in prompt learning for the diagnosis of neurological disorders.

To address the aforementioned challenges, we propose \ours: \emph{Multimodal Medical Data Analysis with Graph Prompt Learning}.
The proposed method is a prompt-driven multimodal medical model for the diagnosis of neurological disorders.
As shown in Fig. \ref{fig_framework}, the proposed \ours{} consists three modules, \ie multimodal data tokenizer (Sec. \ref{sec_tokenizer}), concept learning (Sec. \ref{sec_Concept}), and graph prompt learning (Sec. \ref{sec_Graph}). Firstly, we employ multimodal data tokenizer which projects all modalities into a shared token space. This allows \ours{} to efficiently handle multimodal medical data. 
Secondly, we leverage a general artificial intelligence model (GPT-4) \citep{openai2023gpt} to automatically obtain disease-related concepts \citep{koh2020concept} and further embed these concepts. Additionally, each token (patch embedding) will be compared to all concepts for similarity, and the weight of patch will be calculated based on the token's similarity to its corresponding category concepts. As a result, it addresses the first challenge by handling irrelevant patches. Thirdly, we learn the graph structure among patches based on their relationships with concepts. As a result, it tackles the second challenge by inputting the learned graph structure and the embeddings of corresponding tokens as inputs to a Graph Convolutional Network (GCN) \citep{kipf2017semi}. This allows for the propagation of structural information among patches, enhancing the overall understanding of the connection between different brain regions. To this end, the new embeddings of tokens are fed into the pre-trained unified encoder to obtain the representation of each subject.

Compared to previous methods, the main contributions of our method are summarized as follows.
\begin{itemize}
\item We introduce a novel prompt learning method that effectively reduces the impact of irrelevant patches, which is usually overlooked in existing methods but is crucial in medical data analysis. 
\item Our method innovatively employs a graph prompt to extract structural information among patches. To the best of our knowledge, this is the first attempt to design prompt learning methods with a focus on the pathogenesis of neurological diseases, thereby bridging the gap between multimodal large models and neurological disease diagnosis. Further, \ours~ demonstrates effectiveness across multiple neurological disease datasets and exhibits excellent scalability and flexibility, making it a promising solution for handling multimodal data in the diagnosis of neurological diseases.
\end{itemize}

\section{Related works} \label{sec_related}

\subsection{Multimodal large models}

Unlike traditional large language models such as GPT-3 \citep{brown2020language} and BERT \citep{devlin2018bert}, which can only deal with text data, multimodal large models \citep{radford2021learning, Frozen, LLaVA, PaLM} can integrate multiple modalities, such as language, vision, and audio, to perform a broad spectrum of tasks. 
For instance, CLIP \citep{radford2021learning} leverages a contrastive learning approach for joint pre-training on image-text pairs data.
Besides, LLaVA \citep{LLaVA} conducts image-text representation alignment followed by instruction tuning.
Frozen \citep{Frozen} firstly proposed an approach that employs large language models in multimodal in-context learning. 
Similarly, Video-ChatGPT \citep{maaz2023video} merges a video-adapted visual encoder with a large language model to handle video data.
Recent works like Imagebind \citep{girdhar2023imagebind} and Meta-transformer \citep{zhang2023meta} can learn a joint embedding across more than six different modalities.

In the medical domain, both large language models and multimodal large models have been recently explored for a wide range of medical tasks, including medical question answering and segmentation.
For example, LLaVA-Med \citep{li2023llava} trains a vision-language conversational assistant research questions about medical images.
MedSAM \citep{ma2023segment} fine-tune the segment anything model (SAM) \citep{kirillov2023segment} on medical image data for segmentation task. 
Moreover, Med-PaLM \citep{tu2023towards} and RedFM \citep{wu2023towards} interpret multimodal biomedical data and handle a diverse range of tasks, moving closer to generalist medical artificial intelligence models.
To incorporate a large-scale multimodal dataset, \cite{zheng2023large} build up a large-scale diagnostic dataset comprising 39,026 cases and 192,675 medical images. They utilized this dataset to train a powerful multimodal medical foundation model.
To ensure the trustworthiness of large models, \cite{sun2024trustllm} introduce principles for different dimensions of trustworthiness.
However, due to the complexity of multimodal neuroimaging data, there is still a lack of research and remain challenges in leveraging multimodal large models for diagnosing neurological disorders.

\subsection{Prompt learning}

Prompt learning has emerged as a new paradigm in fine-turning pre-trained large models for various downstream tasks. Compared to full fine-tuning, it achieves comparable performance without updating all parameters in large models. 
In the field of NLP, GPT-3 \citep{brown2020language} has demonstrated strong generalization to downstream tasks by manually selecting prompt texts. To improve the construction of prompt texts,  Prefix-tuning \citep{li2021prefix} optimizes prompts through gradient-based fine-tuning.
In computer vision,  VPT \citep{jia2022visual} prompts the image for effectively using the pre-trained image encoder in the classification task. SAM \citep{kirillov2023segment} takes visual prompts by drawing points, boxes, and strokes on an image for image segmentation. Moreover, MaPLe \citep{khattak2023maple} utilizes multi-modal prompts to guide models in both image and text modalities.
Besides, \cite{miyai2024locoop} apply prompt learning for few-shot out-of-distribution detection.

In the field of medicine,  prompt learning aims to guide models by focusing their attention on relevant regions within complex medical datasets.
For example, several researchers \citep{ma2023segment, huang2023segment, zhang2023input} have introduced prompts such as bounding boxes and clicks into the SAM model to adapt it for medical image segmentation tasks. However, these methods may not be suitable for diagnosing neurological disorders given that they rely on human interaction or object detection models, as well as they do not take into account the importance of brain connectivity in neurological disorders.

\subsection{Graph neural network}
In the past decade, Graph Neural Networks (GNNs) have been widely used in computer-aided diagnosis \citep{sun2020disease, bessadok2022graph, holzinger2021towards, li2021braingnn}. 
GNNs such as Graph Convolutional Network (GCN) \citep{kipf2017semi} and Graph Attention Network (GAT) \citep{velivckovic2017graph} leverage message passing mechanism to capture relationships between nodes and structure information in the graph.
Based on this, GNNs have the capability to leverage the inherent relationships between brain regions to uncover patterns or biomarkers associated with neurological disorder. As a result, there has been a growing interest in applying  GNNs in the field of neuroscience \citep{li2021braingnn}. 
For example, \cite{parisot2018disease} exploits the GCN for Alzheimer’s disease diagnosis in a population graph where the nodes in the graph denote the subjects and edges represent the similarity between subjects. 
Similarly, \cite{kazi2019inceptiongcn} proposes a GCN model with multiple filter kernel sizes on a population graph. Besides, \cite{xu2023graphprompt} leveraged graph structure information for biomedical synonym prediction.

In addition, brain graph is also frequently used in the diagnosis of neurological disorders, where the nodes correspond to anatomical brain regions and the edges depict functional or structural connections between these regions. For instance, \cite{li2021braingnn} conduct an interpretable model on brain graph to determine the specific brain regions that are associated with a particular neurological disorder.
\cite{cai2022graph} propose a transformer-based geometric learning approach to handle multimodal brain graphs for brain age estimation. However, there is still a lack of research on how to effectively leverage graph in multimodal large models.


\section{Methods} \label{sec_Methods}

\subsection{Preliminary and motivations} \label{sec_Motivations}
Utilizing transformers \citep{vaswani2017attention} as the architecture of encoders to process multimodal data has become a popular choice in modern multimodal large models, as it can effectively integrate information from multiple modalities. 
For example, pre-trained vision-language models like CLIP \citep{radford2021learning} employ separate transformer-based backbones (\eg ViT) to encode images and text separately. To obtain representations of the samples, the transformer architecture involves two key components: (i) Tokenization: converting the raw data into tokens.  (ii) Encoding: performing attention-based feature extraction layers on all tokens.

\noindent \textbf{Tokenization}. The raw data from each modality is first tokenized into a sequence of tokens. Given an image input $\mathbf{X}$, the raw pixel data is typically partied into a set of patches $\mathbf{X} = \{\mathbf{x}_1, \mathbf{x}_2, \cdots, \mathbf{x}_m\}$ where $m$ is the number of patches, and each patch is flattened and projected into a token embedding $\mathbf{P} = \{\mathbf{p}_1, \mathbf{p}_2, \cdots, \mathbf{p}_m\}$, which can be formalized as follows:
\begin{equation}\label{Token}
\mathbf{p}_i = \mathcal{F}(\mathbf{x}_i) + \mathbf{E}_{i}^{\text{pos}}, ~~\mathbf{p}_i \in  \mathbb{R}^{D}
\end{equation}
where $\mathcal{F}$ is a learnable projection function, $D$ the dimensionality of token embedding, and $\mathbf{E}_{i}^{\text{pos}}$ is a positional embedding vector that encodes the location of $i$-th patch within the image.

\noindent \textbf{Encoding}. The next step is to process the sequence of token embeddings through the Transformer's encoding layers. Each encoding layer consists of a multi-head self-attention (MHSA) layer followed by an MLP block. The encoding process in  $l$-th layer can be represented as:
\begin{equation}
\begin{aligned}
\mathbf{H}^{(l)} = \text{MHSA}(\text{LN}(\mathbf{Z}^{(l-1)})) + \mathbf{Z}^{(l-1)} \\
\mathbf{Z}^{(l)} = \text{MLP}(\text{LN}(\mathbf{H}^{(l)})) + \mathbf{H}^{(l)},
\label{eq_Encoding}
\end{aligned}
\end{equation}
where $\text{LN}(\cdot)$ is layer normalization operation, $\mathbf{Z}^{(l-1)}$ is the output from the previous layer and the input token embeddings for $l=1$. In the MHSA layer, the attention mechanism computes multiple attention scores by performing dot products between the Query and Key vectors of the tokens, which represent dense interactions among tokens. A single-head attention operation can be expressed as follows:
\begin{equation}
\text{Attention}(\textbf{Q}, \textbf{K}, \textbf{V}) = \text{softmax}\left(\frac{\textbf{Q}\textbf{K}^T}{\sqrt{d_k}}\right)\textbf{V},
\label{eq_Attention}
\end{equation}
where $\textbf{Q}$, $\textbf{K}$, and $\textbf{V}$ are the query, key, and value vectors, respectively, and $d_k$ is the dimensionality of the Key vectors, used for scaling. The softmax function is applied row-wise.

\noindent \textbf{Prompt.} Prompts can be used to adapt the pre-trained multimodal large models for various tasks without fully fine-tuning the model's parameters, which can be constructed through manual design or parameter learning \citep{brown2020language, gao2020making}. The prompt embeddings $\mathbf{P}_{\text{prompt}}$ are integrated with the input token embeddings $\mathbf{P}$:
\begin{equation}
\mathbf{P}_{\text{combined}} = \mathcal{F}_\text{combine} (\mathbf{P}_{\text{prompt}},\mathbf{P}),
\label{eq_combined}
\end{equation}
where $\mathcal{F}_\text{combine}$ denotes a fusion layer that can be a non-parametric operator ($\oplus$ or $+$) or parameterized layers. The combined token embeddings $\mathbf{P}_{\text{combined}}$ are subsequently processed by the transformer encoder, as delineated in Eq. \ref{eq_Encoding}. Based on this, previous methods proposed cross-modality prompt learning \citep{khattak2023maple} and deep prompt learning \citep{liu2022p} to further improve its performance on specific tasks

However, previous methods face two challenges when applied to multimodal neural data: (i). Existing prompt learning methods usually overlook the impact of irrelevant patches. In the study of neural disorders, only a few patches in neural images are pertinent to the disease, which means the majority of tokens in $\{\mathbf{p}_1, \mathbf{p}_2, \cdots, \mathbf{p}_m\}$ represent background information. However, all tokens are treated equally and interact with each other according to Eq. \ref{eq_Encoding} and Eq. \ref{eq_Attention}, thereby limiting their effectiveness.
(ii). Existing prompt learning methods have not taken into account the structural information among patches as indicated by Eq. \ref{eq_combined}. In the field of neuroscience, the intricate network structure of the brain plays a fundamental role in understanding neurological conditions. This highlights the significance of developing advanced methods that can effectively extract and utilize structural information for the diagnosis of neurological disorders.  As a result, the performance of previous methods in the diagnosis of neurological disorders remains suboptimal.

In this paper, we propose a new prompt learning method to address the aforementioned challenges, and the framework is shown in Fig. \ref{fig_framework}. Specifically, we first perform a multimodal data tokenizer in Section \ref{sec_tokenizer} to project the raw data from different modalities into a shard token space, and then design the concept learning for tokens in Section \ref{sec_Concept} and the graph prompt learning in Section \ref{sec_Graph} to address the above issues.


\subsection{Multimodal data tokenizer} \label{sec_tokenizer}
Medical data is inherently multimodal, usually including modalities like MRI, PET, and FC in neurological disorder diagnosis. These diverse modalities provide complementary information, helping to gain a more comprehensive understanding and better analyze neurological disorders. However, these modalities often exhibit more complex data structures compared to natural images, such as 3D tensor medical data, brain connectivity graphs data, and time series data. 
Previous multimodal models like CLIP \citep{radford2021learning} handle text and images by employing distinct tokenizers and encoders for each modality. However, those methods may encounter challenges in terms of efficiency and scalability when dealing with multimodal medical data. 
Additionally, maintaining separate tokenizers and encoders for each modality is inflexible. 
Inspired by Meta-transformer \citep{zhang2023meta} and Imagebind \citep{girdhar2023imagebind}, we employ a multimodal data tokenizer that converts various multimodal medical data into token embeddings. Due to extensive research on the conversion of 2D image data (\eg X-ray and CT) and text data into tokens, there has been relatively less focus on tokenizing 3D tensor medical data (\eg MRI and PET). Therefore, in the following section, we focus on transforming 3D tensor medical data into tokens.

\subsubsection{Patch partitioning}
Specifically, for 3D tensor medical data (\eg MRI and PET),
let us denote the raw data from $M$ modalities as $\mathbf{X} = \{\mathbf{X}^1, \mathbf{X}^2, \ldots, \mathbf{X}^M\}$, where each $\mathbf{X}^m \in \mathbb{R}^{H_m \times W_m \times D_m \times C_m}$ represents a distinct modality with its respective height $H_m$, width $W_m$, depth $D_m$, and number of channels $C_m$. For each modality $m$, we initiate the tokenization process by dividing the data into a set of patches $\mathcal{P}^m = \{\mathbf{p}_1^m, \mathbf{p}_2^m, \ldots, \mathbf{p}_{N_m}^m\}$. 
where $\mathbf{p}_i^m \in \mathbb{R}^{S^3 \cdot C_m}$ is the $i$-th patch, $S$ is the uniform size of each patch in all dimensions, and $N_m = \frac{H_mW_mD_m}{S^3}$ is the total number of patches for $m$+th modality. 

There are multiple approaches to dividing the data into a set of patches, including (i) 2D slice patch: each volumetric scan is sliced along one dimension, and each 2D slice is then partied into patches of size $S \times S$. The advantage of this approach lies in the fact that multimodal large models are typically pre-trained on 2D images, making it easier to transfer them to 2D medical images. However, this approach will generate a large number of tokens, which increases the computational and makes it more challenging to optimize.
(ii) 2D axial slice patch: slices are taken along the axial plane, which is parallel to the plane of the 3D tensor's horizon, and each slice is divided into patches of size $S \times S$. The advantage of this approach is that it generates a considerable number of tokens. However, one drawback is that it can result in information loss. (iii) 3D patch: the 3D tensor data is divided into smaller cubes, with each cube being of size $S \times S \times S$. The advantage of this approach is that it generates a considerable number of tokens without causing information loss. However, one drawback is that it may contain gaps between tokens and the pre-training fundamental models. Note that, each approach has its advantages and disadvantages. Therefore, it is necessary to choose the most suitable method based on the specific task and characteristics of the data.

\subsubsection{Tokenization}
Each patch $\mathbf{p}_i^m$ is then transformed into a token through a modality-specific patch projection layer $\mathcal{F}_{\text{token}}^m$ and align the dimensions of the token embeddings from each modality, which can be expressed as follows:

\begin{equation}
\mathbf{t}_i^m = \mathcal{F}_{\text{token}}^m(\mathbf{p}_i^m) + \mathbf{E}_{i}^{m,\text{pos}},
\label{eq_token}
\end{equation}
where $\mathbf{E}_{i}^{m,\text{pos}}$ is the position embedding of each patch in $m$-th modality and the modality-specific patch projection layer $\mathcal{F}_{\text{token}}^m$ maps each patch $\mathbf{p}_i^m$ into a $D$-dimensional token embedding $\mathbf{t}_i^m$. As a result, we obtain all tokens from different modalities, \ie $\mathbf{T}^m = \{\mathbf{t}_1^m, \mathbf{t}_2^m, \ldots, \mathbf{t}_{N_m}^m\}$.

We further project token embedding from multiple modalities into a shared token embedding space by a common learnable liner projection layer, which can be formalized as:
\begin{equation}
\mathbf{t}_i^m = \mathcal{F}_{\text{align}}(\mathbf{t}_i^m) + \mathbf{E}_{\text{modality}}^{m},
\label{eq_align}
\end{equation}
where $\mathbf{E}_{\text{modality}}^{m}$ is the position embedding of each modality and $\mathcal{F}_{\text{align}}$ a common learnable liner projection layer. As a result, we obtain all tokens from all modalities, \ie $\mathbf{T} = \{\mathbf{t}_1, \mathbf{t}_2, \ldots, \mathbf{t}_{N}\}$.

\subsection{Concept learning} \label{sec_Concept}
Once all token embeddings are obtained, it is crucial to consider the importance of each token because only a small fraction of tokens are related to the disease. However, it is challenging to identify these disease-related tokens due to the lack of annotations and the high dimensionality of tokens. 
To address this challenge, we propose to utilize concepts \citep{koh2020concept} of disease and compute the semantic similarity between each token and all concepts. We further reduce the weight or importance of irrelevant tokens according to the semantic similarity between each token and disease-related concepts.

\subsubsection{Concept generation} \label{sec_Concept_gen}
Instead of providing low-level semantic information from the patches, the concepts refer to abstracting meaningful patterns from the data \citep{wang2023learning}, thereby offering higher-level information that connects the patches to specific categories. In neurological disorders diagnosis, concepts are usually related to disease-specific information, such as symptoms, biomarkers, or radiological characteristics.

A straightforward approach to generating a set of concepts for diagnosing neurological disorders is to leverage human expertise. Handcrafting a set of concepts offers better interpretability as they align with human perception and understanding \citep{wang2023learning}. However, it is important to note that the process of annotating these concepts can be costly, as well as requiring rich medical expertise. To avoid those issues, previous methods attempted to use GPT-4 \citep{openai2023gpt} or other large language modal to generate concept texts. For example, \citep{yang2023language} proposed to prompt the large language model (\ie GPT-3) to generate candidate concepts. Based on this, \citep{yan2023robust} proposed to elicit medical knowledge from GPT-4 to build a set of concepts for interpretable medical image classifiers. Thus, building concepts from large language models (\eg GPT-3 and GPT-4) has been success applied. In this study, we use GPT-4 to automatically generate the diseases-related concepts. Specifically, by prompting GPT-4 with specific instructions, we can generate lists of concepts that are typically associated with diseases, \eg ``\textit{Reduced Brain Metabolism: Individuals with AD typically exhibit slowed metabolic activity in the brain, particularly in the frontal and temporal lobes}''. This approach reduces the annotation cost and leverages the capabilities of multimodal large models to generate concepts based on vast amounts of medical knowledge.
More specifically, for $C$ categories in specific neurological disorder diagnoses, we generate $K$ relevant concepts for each category, which can be formalized as:
\begin{equation}
\mathbf{X}_{\text{cpt}} = \mathbf{GPT}(\{\text{CLS}_1, \text{CLS}_2, \ldots, \text{CLS}_C\},K),
\label{eq_GPT}
\end{equation}
where $\mathbf{GPT}$ denotes GPT-4 API \citep{openai2023gpt}, $\text{CLS}$ denotes the name of corresponding category, and $\mathbf{X}_{\text{cpt}}$ denotes the generated text corresponding to $C \times K$ concepts. In this study, the concept texts have been validated by clinicians to ensure they are correct.

\subsubsection{Semantic similarity computation}
The obtained concepts are utilized to calculate the semantic similarity between each token $t_{i}$, and the disease-related concepts. This process assists in identifying the most pertinent tokens and adjusting their weights accordingly. As a result, it prompts the model to focus on important tokens, reducing interference from irrelevant tokens. To achieve this, we first input the text of concepts $\mathbf{X}_{\text{cpt}}$ with its corresponding category name into the text encoder $\mathcal{F}_{\text{text}}(\cdot)$ to obtain embeddings of concepts $\mathbf{Z}_{\text{cpt}}$, \ie
\begin{equation}
\mathbf{Z}_{\text{cpt}}^{c,k} = \mathcal{F}_{\text{text}}(\{\mathbf{X}_{\text{cpt}}^{c,k}, \text{CLS}_c\}),
\label{eq_textembedding}
\end{equation}
where $\mathbf{Z}_{\text{cpt}}^{c,k}$ denote the final embedding of $k$-th concept in $c$-th category. Note that, the text encoder $\mathcal{F}_{\text{text}}(\cdot)$ and the tokenizer encoder have not pre-trained on paired data, so the distributions of $\mathbf{Z}_{\text{cpt}}$ and $\mathbf{T}$ are not aligned. Thus, we apply a learnable projection layer $\mathcal{F}_{\text{pro}}(\cdot)$ on tokens $\mathbf{T}$ to align their distributions. Next, we calculate the semantic similarity between tokens and concepts, \ie
\begin{equation}
\mathbf{s}_{i,c} = \frac{\text{exp}(\text{sim}(\mathcal{F}_{\text{pro}}(\mathbf{t}_i),  \mathbf{Z}_{\text{cpt}}^{c})/ \tau )}
{ {\textstyle \sum_{j=1}^{C \times K}} \text{exp}(\text{sim}(\mathcal{F}_{\text{pro}}(\mathbf{t}_i), \mathbf{Z}_{\text{cpt}}^{j})/ \tau )},
\label{eq_similarity}
\end{equation}
where $\text{sim}(\cdot, \cdot)$ denotes the cosine similarity operator and $\tau$ denotes temperature parameter. After that, we calculate the weights of the tokens based on their relevance to the category-related concepts by:
\begin{equation}
w_{i} = \frac{{\textstyle \sum_{j \in \mathcal{C}_i}} \mathbf{s}_{i,j}}
{ {\textstyle \sum_{j=1}^{C \times K}} \mathbf{s}_{i,j}} \times C,
\label{eq_similarity} 
\end{equation}
where $\mathcal{C}_i$ is the set of concepts belonging to the category of this subject. Finally, we adjust the weight of the tokens and obtain the weight token embedding, \ie $\tilde{t_{i}} = w_{i}t_{i}$.
Note that, as the category of the sample is unknown, we select  $\mathcal{C}_i$ belongs to which set of concepts with the highest weight during the inference.

Finally, we adjust the weights of the tokens based on their relevance to the category-related concepts. Tokens with higher weights are deemed more pertinent in subsequent processing. This process helps the model to focus on the most relevant tokens and reduces the noise from irrelevant tokens.
In addition, it is also necessary to consider the pathogenesis and biomarkers of neurological disorders when designing prompts.

\subsection{Graph prompt learning} \label{sec_Graph}
Neuroscience researchers \citep{rubinov2010complex, belmonte2004autism} have elucidated that the brain constitutes a complex graph structure, comprised of brain regions. The structural information of this graph, specifically the pattern of its connections, is crucial in the pathogenesis of neurological disorders. For example, deterioration and abnormal connectivity are candidate biomarkers for Alzheimer's disease  \citep{pievani2014brain}. 
Hence, the implementation of graph prompts is both reasonable and necessary for improving the ability of pre-trained multimodal large models to tackle the diagnosis of neurological disorders.
To achieve this, we first construct the graph, and then extract embeddings from the constructed graph.

\subsubsection{Graph construction}
Considering that each token represents embedding of its corresponding patch (\ie local brain region), 
we first regard the tokens as nodes in the graph. Then, we construct the edges/graph structural $\mathbf{A}$ based on semantic relationships between tokens. 
Notably, directly calculating the relationships between tokens based on token embeddings $\mathbf{T}$ may not be the optimal choice since token embeddings $\mathbf{T}$ may contain limited information about disease-related information. To solve this issue, we leverage the concept embeddings obtained in Sec. \ref{sec_Concept} as the bridge to learn the connections between tokens. With the the semantic similarity between tokens and concepts $\mathbf{S}$, the connection $\mathbf{a}_{i,j}$ between $i$-th token and $j$-th token is calculated by:
\begin{equation}
\mathbf{a}_{i,j} = \frac{\text{exp}(\text{sim}(\mathbf{S}_{i},  \mathbf{S}_{j})/ \tau )}
{ {\textstyle \sum_{j=1}^{N}} \text{exp}(\text{sim}(\mathbf{S}_{i},  \mathbf{S}_{j})/ \tau )},
\label{eq_graph}
\end{equation}
where $\mathbf{S}_{i}$ is the semantic similarity between $i$-th token and all concepts, and $\tau$ denote temperature parameter.
Intuitively, tokens belonging to similar concepts are more likely to be connected with a higher probability.
This approach offers two advantages compared to directly calculating connection probabilities based on token embeddings. Firstly, it mitigates potential noise connections caused by irrelevant features in high-dimensional embedding $\mathbf{t}$. Secondly, the graph structure constructed based on concepts related to biomarkers and radiological characteristics of the disease is more meaningful in neuroscience.
Based on this, we treat the constructed graph structure $\mathbf{A}$ as the prompt for pre-trained foundation modal. 

\subsubsection{Graph embedding}

After obtaining the constructed graph structure $\mathbf{A}$, we employ the widely-adopted GCN model as the graph encoder to obtain the graph embedding in this study. The GCN operation in $l$-th GCN layer is formally defined as:
\begin{equation}
\mathcal{F}_{\text{GCN}}(\mathbf{A}, \mathbf{H}) = \sigma(\tilde{\mathbf{D}}^{-\frac{1}{2}}\tilde{\mathbf{A}}\tilde{\mathbf{D}}^{-\frac{1}{2}}\mathbf{H}\mathbf{\Theta}),
\label{eq:gcn}
\end{equation}
where $\tilde{\mathbf{A}} = \mathbf{A} + \mathbf{I}$ is the adjacency matrix $\mathbf{A}$ with added self-connections through the identity matrix $\mathbf{I}$, $\tilde{\mathbf{D}}$ is the diagonal matrix of $\tilde{\mathbf{A}}$, $\mathbf{H}^{(l)}$ is the input embeddings of all nodes, $\mathbf{\Theta}$ is the trainable weight matrix, and $\sigma(\cdot)$ represents an activation function. Furthermore, the embeddings of all tokens with graph prompt can be expressed as:
\begin{equation}
\mathbf{T}^{G} = \mathcal{F}_{\text{GCN}}(\mathbf{A}, \tilde {\mathbf{T}}),
\label{eq:TG}
\end{equation}
where $\tilde{\mathbf{T}}$ is the weighted embeddings of the tokens. As a result, our method obtains the prompted token embeddings $\mathbf{T}^{G}$ by extracting the structural information among tokens to prompt the pre-trained fundation models. 
Furthermore, the prompted token embeddings $\mathbf{T}^{G}$ are fed into a unified transformer-based encoder $\mathcal{F}_{uni}(\cdot)$ to obtain the representation of the subject, \ie $\mathbf{z} = \mathcal{F}_{uni}(\mathbf{T}^{G}), \mathbf{z} \in \mathbb{R}^{d}$. 

Finally, we produce a prediction of the subject by two functions, \ie concept project function $\phi : \mathbb{R}^{d} \to  \mathbb{R}^{K \times C}$ and label project function $\mathcal{F}_y: \mathbb{R}^{K \times C} \to  \mathcal{Y}$.
 In this way,  we apply cross-entropy loss as the objective function, \ie
\begin{equation}
\mathcal{L}_{\text{CE}} = -\frac{1}{N_l}\sum_{i=1}^{N_l} 
 \mathbf{y}_{i} \log \left(\mathcal{F}_y\left(\phi \left(\mathbf{z}_{i}, \mathbf{Z}_{cpt}\right)\right)\right),
\label{eq:loss}
\end{equation}
where $\mathbf{y} \in \mathcal{Y}$ denotes label, $\mathbf{Z}_{cpt}$ is the concept embeddings obtained from Eq. (\ref{eq_textembedding}), and $N_l$ is number of labeled subjects.

\section{Experiments}
\subsection{Experimental settings}
\subsubsection{Datasets}
We conduct experiments on two public multimodal neurological disorder datasets, \ie Alzheimer's Disease Neuroimaging Initiative (ADNI)\footnote{https://adni.loni.usc.edu/} \citep{jack2008alzheimer} and Autism brain imaging data exchange (ABIDE)\footnote{http://fcon\_1000.projects.nitrc.org/indi/abide/} \citep{di2014autism}.
The subjects are categorized into four groups in ADNI:  NC (normal control), EMCI (early mild cognitive impairment), LMCI (late mild cognitive impairment), and AD (Alzheimer’s disease). In total, we included 409 pairs of data (\ie NC/LMCI/AD for ADNI-3CLS) and 770 pairs of data (\ie NC/EMCI/LMCI/AD for ADNI-4CLS). The subjects are categorized into two groups in ABIDE: NC (normal control) and ASD (autism spectrum disorder) and we included 1029 pairs of data in ABIDE dataset.

We utilized the fslreorient2std, robustfov, FLIRT, and BET tools from the FSL software \citep{jenkinson2012fsl} to preprocess data, the pipeline are follows. We first employed the fslreorient2std tool to reorient the images, aligning them with the orientation of the standard template image. Next, we utilized the robustfov tool to crop the MRI images, effectively removing the neck and lower jaw regions. The FLIRT tool \citep{jenkinson2002improved} was used to register all MRI images to the Colin27 template \citep{holmes1998enhancement}, correcting for global linear differences and resampling the images to a consistent resolution (1 × 1 × 1mm³) and size (191 × 217 × 191). Finally, we applied the BET tool to remove the skull and dura mater.

\begin{table*}[!ht]
\small
\centering
\caption{Diagnose Performance (mean and standard deviation) of all methods on all datasets. Note that, ``ADNI-3CLS'' and ``ADNI-4CLS'' indicate the classification on three classes ``CN/LMCI/AD'' and the classification on four classes ``CN/EMCI/LMCI/AD'', respectively.}
\renewcommand\tabcolsep{6pt}
\resizebox{\textwidth}{!}{
\begin{tabular}{l@{\hspace{1cm}}c@{\hspace{1cm}}c@{\hspace{1cm}}c@{\hspace{1cm}}c@{\hspace{1cm}}c}

\specialrule{2pt}{2pt}{2pt}

\multirow{2}{*}{Methods}&\textbf{ADNI-3CLS}&&&\\
\cline{2-6} 
 & ACC & AUC &   SPE & SEN & F1 \\ 
\hline
ViT     &0.6707 $\pm$ 0.0387  &0.6997 $\pm$ 0.0411  &0.5837 $\pm$ 0.0332  &0.5749 $\pm$ 0.0402  &0.5749 $\pm$ 0.0323   \\ 
BrainT  &0.6504 $\pm$ 0.0640  &0.6853 $\pm$ 0.0422  &0.5589 $\pm$ 0.0358  &0.5806 $\pm$ 0.0328  &0.5692 $\pm$ 0.0353   \\ 
GraphT  &0.6911 $\pm$ 0.0744  &0.7172 $\pm$ 0.0810  &0.5729 $\pm$ 0.0580  &0.5954 $\pm$ 0.0602  &0.5757 $\pm$ 0.0440   \\ 
CLIP    &0.7073 $\pm$ 0.0467  &0.7361 $\pm$ 0.0483  &0.5671 $\pm$ 0.0383  &0.6144 $\pm$ 0.0363  &0.5869 $\pm$ 0.0376   \\ 
MetaT   &0.7114 $\pm$ 0.0365  &0.7521 $\pm$ 0.0415  &0.5674 $\pm$ 0.0303  &0.6252 $\pm$ 0.0482  &0.5944 $\pm$ 0.0368   \\ 
VPT     &0.7561 $\pm$ 0.0222  &0.7891 $\pm$ 0.0272  &0.6049 $\pm$ 0.0277  &0.6627 $\pm$ 0.0293  &0.6322 $\pm$ 0.0283   \\ 
MaPLe   &0.7805 $\pm$ 0.0345  &0.8093 $\pm$ 0.0516  &0.7272 $\pm$ 0.0497  &0.6775 $\pm$ 0.0482  &0.6708 $\pm$ 0.0356   \\ 
 \textbf{\ours{}}&0.8230 $\pm$ 0.0310& 0.8514 $\pm$ 0.0172& 0.7714 $\pm$ 0.0126& 0.7314 $\pm$ 0.0412& 0.7467$\pm$ 0.0312 \\ 
\specialrule{2pt}{2pt}{2pt}
\multirow{2}{*}{Methods}&\textbf{ADNI-4CLS}&&&\\
\cline{2-6} 
 & ACC & AUC &   SPE & SEN & F1 \\  
\hline
ViT     &0.3766 $\pm$ 0.0165 &0.4530 $\pm$ 0.0179 &0.3660 $\pm$ 0.0167 &0.3631 $\pm$ 0.0161 &0.3644 $\pm$ 0.0142  \\ 
BrainT  &0.4091 $\pm$ 0.0224 &0.5859 $\pm$ 0.0210 &0.4246 $\pm$ 0.0192 &0.3680 $\pm$ 0.0196 &0.3681 $\pm$ 0.0211  \\ 
GraphT  &0.3961 $\pm$ 0.0409 &0.5188 $\pm$ 0.0373 &0.3961 $\pm$ 0.0429 &0.3388 $\pm$ 0.0387 &0.3143 $\pm$ 0.0431 \\ 
CLIP    &0.4351 $\pm$ 0.0221 &0.6013 $\pm$ 0.0230 &0.4791 $\pm$ 0.0261 &0.4077 $\pm$ 0.0305 &0.4247 $\pm$ 0.0286  \\ 
MetaT   &0.4286 $\pm$ 0.0312 &0.5779 $\pm$ 0.0347 &0.4922 $\pm$ 0.0313 &0.4061 $\pm$ 0.0226 &0.4136 $\pm$ 0.0248  \\ 
VPT     &0.4416 $\pm$ 0.0303 &0.6198 $\pm$ 0.0297 &0.4455 $\pm$ 0.0252 &0.4149 $\pm$ 0.0251 &0.4219 $\pm$ 0.0295  \\ 
MaPLe   &0.4675 $\pm$ 0.0372 &0.6357 $\pm$ 0.0275 &0.4922 $\pm$ 0.0202 &0.4641 $\pm$ 0.0288 &0.4722 $\pm$ 0.0305  \\ 
 \textbf{\ours{}}&0.5159 $\pm$ 0.0184 & 0.6422 $\pm$ 0.0484 & 0.5252 $\pm$ 0.0308 & 0.4317 $\pm$ 0.0243 & 0.4779 $\pm$ 0.0179\\ 
\specialrule{2pt}{2pt}{2pt}
\multirow{2}{*}{Methods}&\textbf{ABIDE}&&&\\
\cline{2-6} 
 & ACC & AUC &   SPE & SEN & F1 \\  
\hline
ViT
&0.6463 $\pm$ 0.0443
&0.6574 $\pm$ 0.0410
&0.5216 $\pm$ 0.0501
&0.5852 $\pm$ 0.0269
&0.5888 $\pm$ 0.0279
\\ 
BrainT
&0.7073 $\pm$ 0.0402
&0.7360 $\pm$ 0.0440
&0.6769 $\pm$ 0.0219
&0.6280 $\pm$ 0.0530
&0.6463 $\pm$ 0.0376
\\ 
GraphT
&0.6585 $\pm$ 0.0215
&0.6911 $\pm$ 0.0424
&0.5507 $\pm$ 0.0372
&0.5340 $\pm$ 0.0285
&0.5523 $\pm$ 0.0279
\\ 
CLIP
&0.6707 $\pm$ 0.0416
&0.6868 $\pm$ 0.0424
&0.6313 $\pm$ 0.0518
&0.5656 $\pm$ 0.0665
&0.6035 $\pm$ 0.0377
  \\ 
MetaT
&0.6585 $\pm$ 0.0228
&0.6880 $\pm$ 0.0196
&0.5737 $\pm$ 0.0472
&0.5675 $\pm$ 0.0505
&0.5761 $\pm$ 0.0465
\\ 
VPT 
&0.6973 $\pm$ 0.0305
&0.7363 $\pm$ 0.0356
&0.6500 $\pm$ 0.0456
&0.6267 $\pm$ 0.0426
&0.6356 $\pm$ 0.0501
 \\ 
MaPLe
&0.7195 $\pm$ 0.0326
&0.7418 $\pm$ 0.0380
&0.6682 $\pm$ 0.0232
&0.6024 $\pm$ 0.0229
&0.6580 $\pm$ 0.0319
\\ 
  \textbf{\ours{}}&0.7239 $\pm$ 0.0229 
& 0.7540 $\pm$ 0.0485 
& 0.6422 $\pm$ 0.0328 
& 0.6897 $\pm$ 0.0272 
& 0.6723 $\pm$ 0.0413  \\ 
\specialrule{2pt}{2pt}{2pt}
\end{tabular}
}
\label{tab:1}
\end{table*}

\subsubsection{Comparison methods}
The comparison methods include one baseline transformer method (\ie ViT \citep{dosovitskiy2020image}), two transformer-based methods for neurological disorders diagnosis (\ie BrainT \citep{kan2022brain} and GraphT \citep{cai2022graph}), two vanilla multi-modal large models (\ie CLIP \citep{radford2021learning} and MetaT \citep{zhang2023meta}), and two prompt learning methods (\ie VPT \citep{jia2022visual} and MaPLe \citep{khattak2023maple}). We list the details of the comparison methods as follows:
\begin{itemize}
\item  \textbf{ViT} is treated as a baseline method. ViT applies the core ideas of transformers to image classification tasks. It treats images as sequences of patches and processes them using self-attention mechanisms, setting a precedent for subsequent transformer-based models in computer vision.

\item  \textbf{BrainT} leverages the transformer to efficiently learn connection strengths between brain regions and incorporate an orthonormal clustering readout operation to capture functional modules within the brain.

\item  \textbf{GraphT} introduces a graph transformer framework that leverages multimodal neuroimaging data to improve Alzheimer's Disease diagnosis and brain age estimation by capturing complex cross-modal interactions and fusing them through geometric learning.


\item  \textbf{CLIP}  is a multi-modal large model, which learns visual concepts from natural language supervision. It has been shown to be effective in a wide range of visual tasks by leveraging the power of language-image pre-training.

\item  \textbf{MetaT}   introduces a unified approach to multimodal learning, employing a frozen encoder and a shared token space to process diverse data types without paired training data.

\item  \textbf{VPT}  introduces efficient prompt learning to full fine-tuning for large-scale transformer models in computer vision.
This groundbreaking methodology not only streamlines
the training process but also significantly optimizes the utilization of computational resources.

\item \textbf{MaPLe} introduces a multimodal prompt learning method that strategically masks parts of the input to guide the model towards more effective learning and adaptation to specific tasks, including those involving multi-modal data.

\end{itemize}

\subsubsection{Implementation details}
All experiments are conducted on a server with  8  NVIDIA  GeForce  3090 GPU (24.0GB caches). 
In our method, we utilize the text encoder and vision encoder from the pre-trained multimodal model BiomedCLIP \citep{zhang2023biomedclip} as the initial weights for our text encoder and unified encoder.  Moreover, BiomedCLIP was trained on 15 million pairs of medical image-text data. By leveraging the pre-trained BERT model from BiomedCLIP, we are able to effectively align concept texts with embeddings that possess semantic meaning in the image space.
To achieve efficient fine-tuning, the parameters of these encoders do not need to be updated (\ie frozen) during the fine-tuning process. Furthermore, this ensures that the embeddings obtained from the text encoder and unified encoder are aligned with each other.
Additionally, all parameters are initialized by the Glorot initialization and optimized by the AdamW optimizer  with initial learning rate $0.0001$, and decay it
by 0.2 at epochs 30 and 60. \ours{} is trained with a fixed epoch 100. The patch size is set in the range of $\{16,24,32,64\}$ and the batch size is set in the range of $\{4,8,16\}$ according to memory.
We obtained the author-verified codes for all comparison methods and tune the parameters for all comparison methods as suggested in their published works. Furthermore, for methods that are originally single-modal, we adopt the results from the best-performing modality as the final outcome. In the case of BrainT, we replace its tokenizer module to adapt to the modality data utilized in the experiment.

We use 5-fold cross-validation in all experiments and repeat the experiments three times with random seeds. Finally, the average results and the corresponding standard deviation results (std) for each method are reported. All methods' results are evaluated using five metrics: Accuracy (ACC), Area Under the Receiver Operating Characteristic Curve (AUC), Specificity (SPE), Sensitivity (SEN), and F1-score (F1). 

\subsection{Experimental results and analysis}

\subsubsection{Diagnose performance}
Table \ref{tab:1} summarize the diagnose performances (mean and standard deviation) of all methods on all datasets. Form the experimental results, we could have the following conclusions.

Firstly, the proposed \ours{} performs comparably or significantly better than the state-of-the-art methods.
For example, compared to one of the best competitors (\eg MaPLe), \ours{} averagely improves by about 7.1\% on ADNI-3CLS, 2.1\% on ADNI-4CLS, and 3.0\% on ABIDE, respectively.
Moreover, compared to one of the baseline competitors (\eg MetaT), \ours{} averagely improves by about 21.4\% on ADNI-3CLS, 12.0\% on ADNI-4CLS, and 13.9\% on ABIDE, respectively.
These results validate the effectiveness of our method. The possible reasons can be that we reduce the weight of irrelevant patches and incorporate a graph prompt to extract structural information among patches, making our method particularly suitable for neurological disorders diagnosis.

\begin{figure*}[!t]
\centering
\subfigure[ADNI-3CLS]{\includegraphics[height=1.4in]{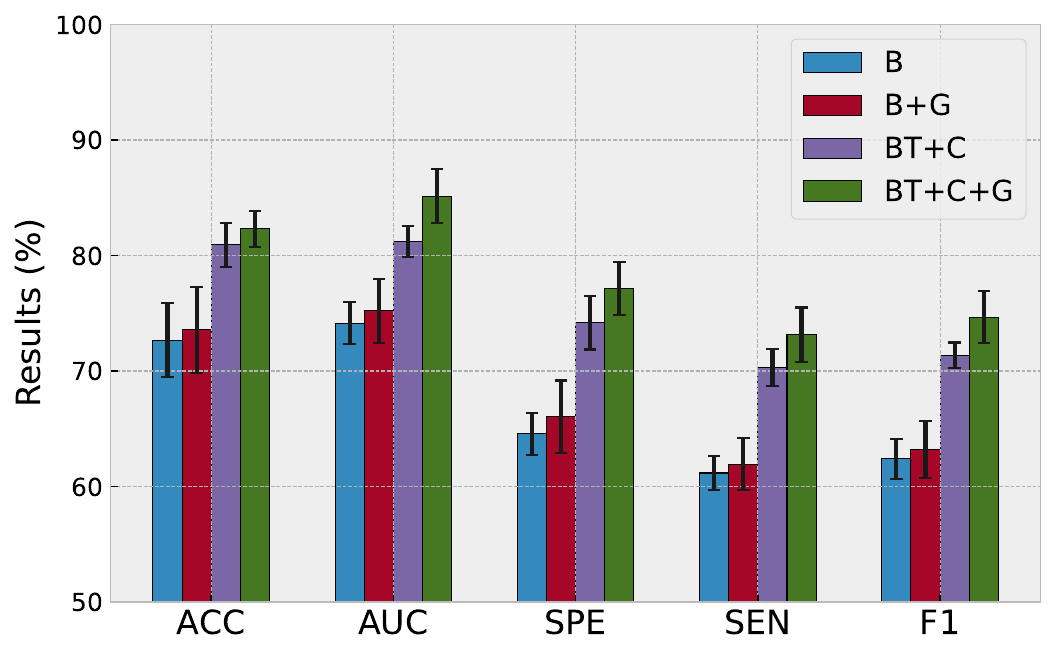}}
\subfigure[ADNI-4CLS]{\includegraphics[height=1.4in]{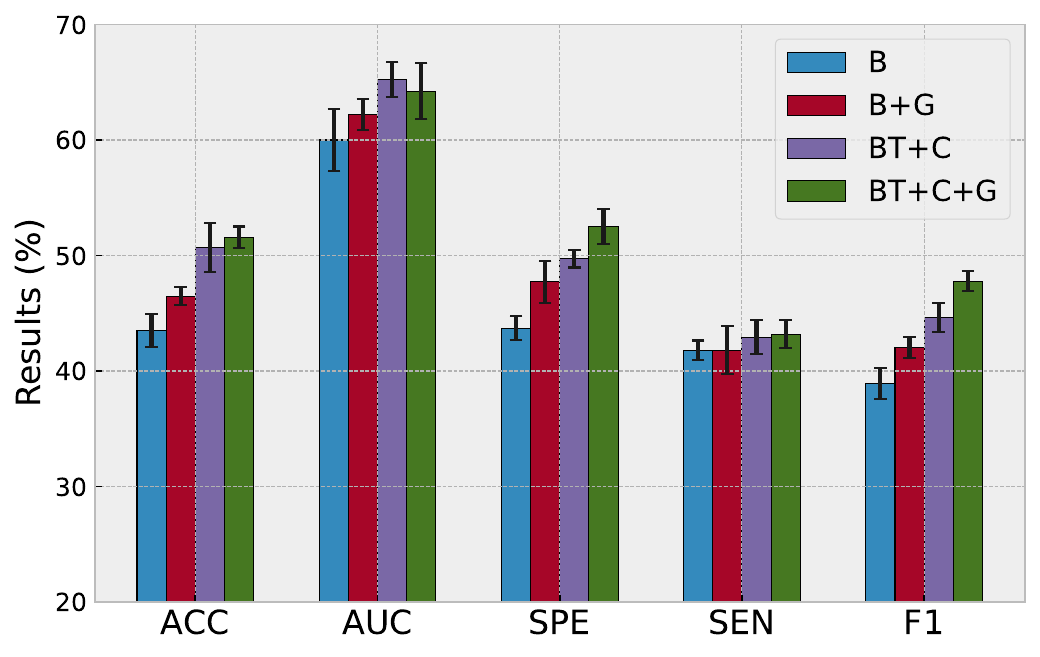}}
\subfigure[ABIDE]{\includegraphics[height=1.4in]{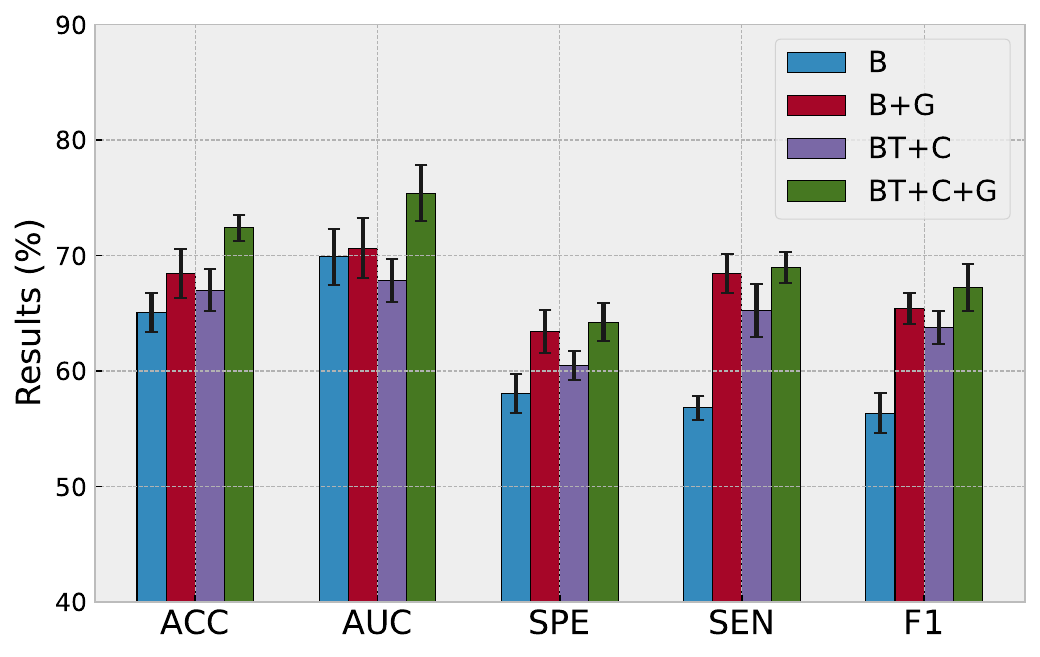}}
\caption{Performance of \ours{} with different combination of components on all datasets, \ie ``B'' denotes baseline method, ``B+G'' denotes baseline method with graph prompt learning, ``B+C'' denotes baseline method with concept learning, and ``B+C+G'' denotes baseline method with graph prompt learning and concept learning.}
\label{fig:ab}
\end{figure*}

Secondly, among all methods, methods that utilize pre-trained large models (\eg CLIP, MetaT, VPT, MaPLe, and \ours{}) outperform other methods trained directly on the dataset (\eg ViT, BrainT, and GraphT). 
For example, compared to ViT, BrainT, and GraphT, \ours{} averagely improves by about 25.2\% on ADNI-3CLS, 31.3\% on ADNI-4CLS, and 16.5\% on ABIDE, respectively.
The superior performance of pre-trained models can be attributed to their exposure to a diverse range of tasks during pre-training, which enhances their ability to generalize to new tasks. This advantage is particularly beneficial when dealing with relatively small target datasets, as is frequently encountered in neurological disorder diagnosis. 

Thirdly, among all methods, methods that utilize prompt learning technique (\eg VPT, MaPLe, and \ours{}) outperform vanilla turning method on pre-trained large models (\eg CLIP and MetaT). 
For example, compared to CLIP and MetaT, \ours{} averagely improves by about 24.5\% on ADNI-3CLS, 11.2\% on ADNI-4CLS, and 9.5\% on ABIDE, respectively.
The possible reason for this is that prompt learning typically involves tuning a small set of parameters that guide the model to apply its pre-learned knowledge in a specific and task-relevant manner.

\begin{figure}[!t]
\centering
{
\scalebox{0.34}{\includegraphics{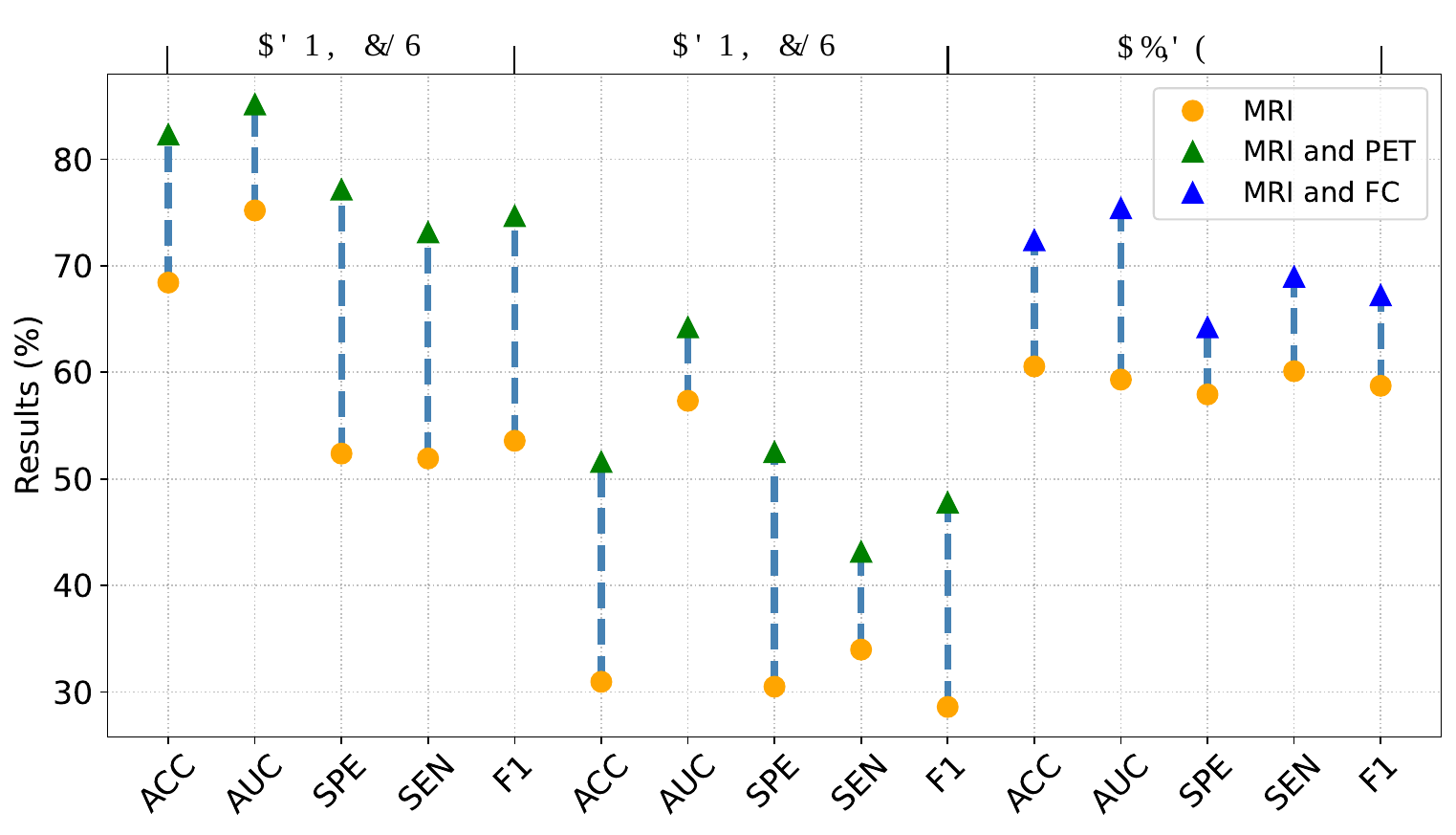}}\\ 
\vspace{1mm}
\scalebox{0.34}{\includegraphics{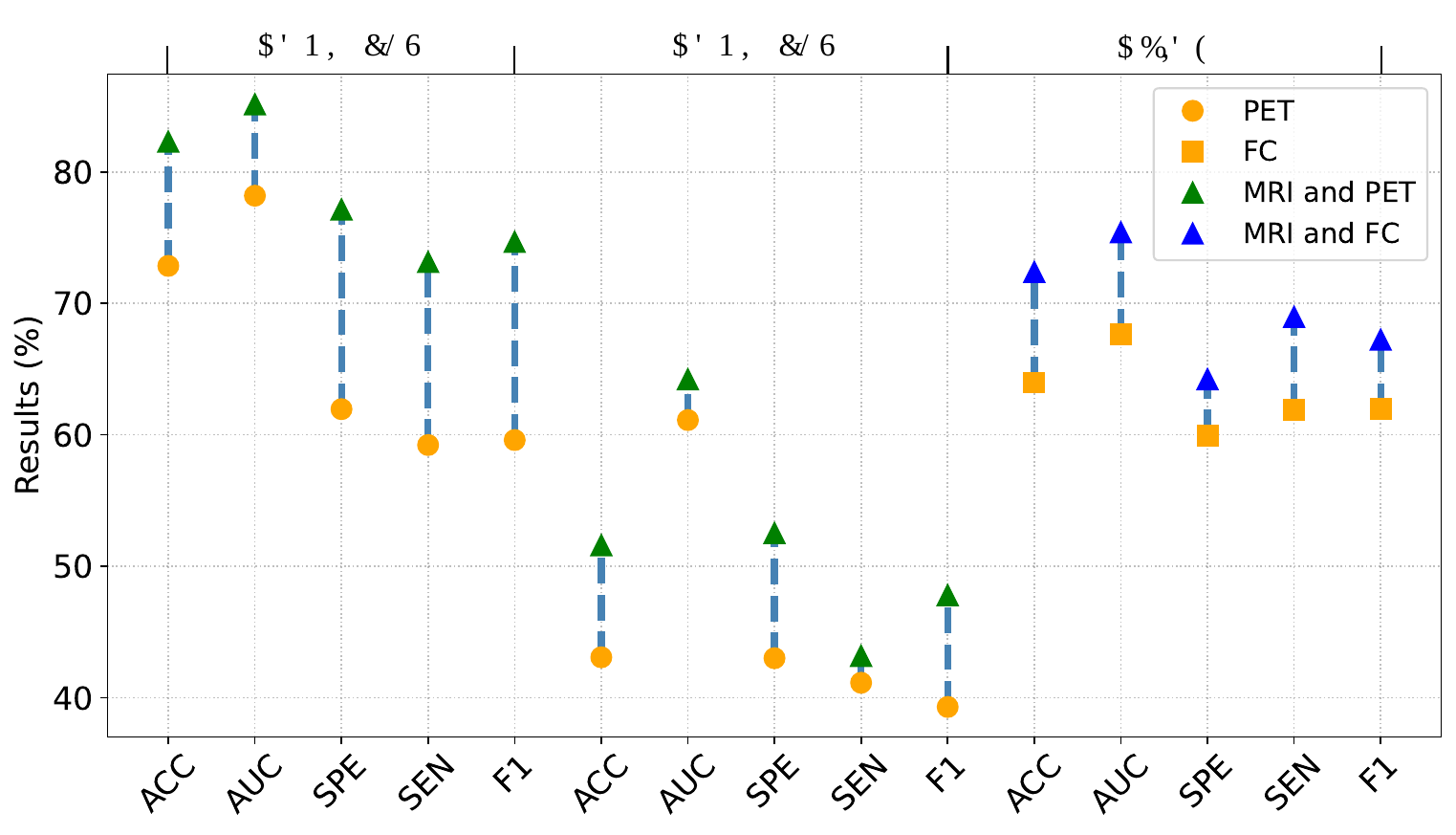}}
}
\caption{Performance of \ours{} with different modalities.}
\label{fig_dim}
\end{figure}

\begin{figure*}[!ht]
\centering
{
\scalebox{1}{\includegraphics{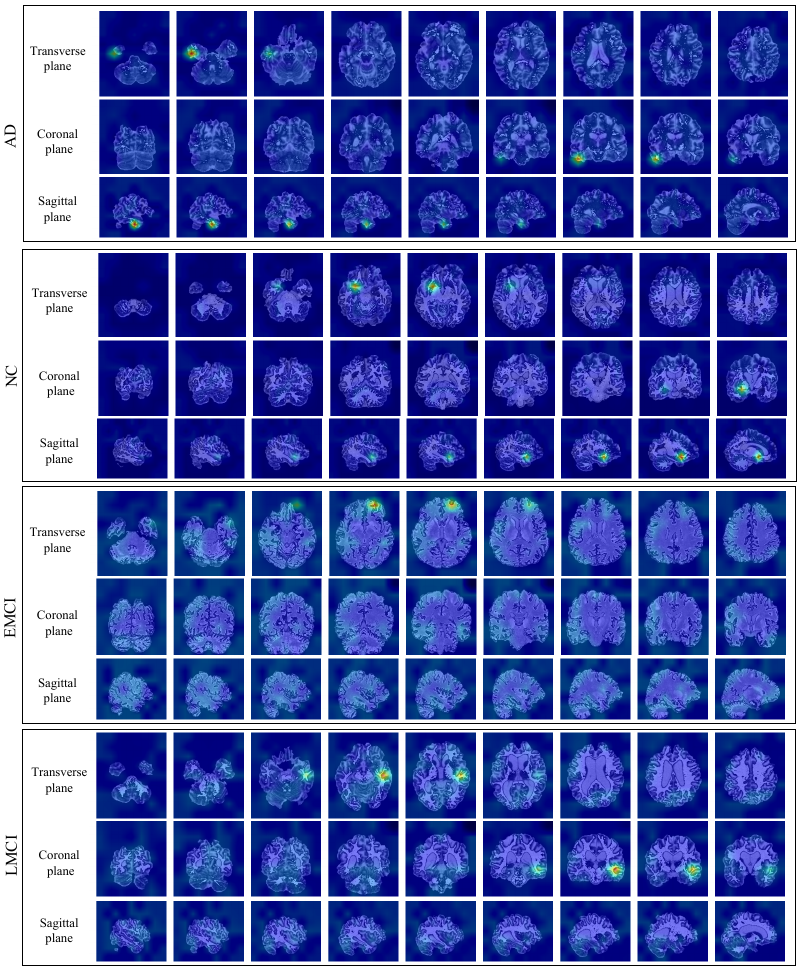}}
}
\vspace{-1mm}
\caption{Heat maps generated by \ours{} on different subjects in ADNI dataset.}
\label{fig_vis}
\end{figure*}

\subsubsection{Ablation studies}
We further conduct the ablation studies with individual components and report their results in Fig. \ref{fig:ab}.
Specifically, ``B'' denotes the baseline method which applies the fine-turning strategy introduced in method MetaT \citep{zhang2023meta}. Based on this, we apply concept learning (\ie Sec. \ref{sec_Concept}) to reduce the weights of irrelevant patches and denote as ``B+W''. Similarly, we employ graph prompt learning (\ie Sec. \ref{sec_Graph}) and denote as ``B+G''. Note that, in order to ensure the proper running of the method, the semantic similarity is still needed to calculate in method ``B+G''. Moreover, we involve both token weights and graph prompt learning which denotes as ``B+W+G''.

Based on the experimental results, we could observe that each component has a contribution. In particular, compared to the baseline method ``B'',  the method ``B+G'' averagely improves by 5.3\%. 
Moreover, the method ``B+W'' markedly outperforms that of the baseline method ``B''.
Hence, the proposed concept learning and graph prompt learning modules have a positive effect on the diagnosis of neurological disorders, which verified our claim that reduces the weight of irrelevant patches and incorporates a graph prompt making our method particularly suitable for neurological disorders diagnosis.
Besides, we observed that concept learning with token weights contributes more significantly to the overall performance compared to graph prompts. One possible reason is that without token weights, the inclusion of many irrelevant tokens during graph prompt learning could have a negative impact. Considering the message propagation mechanism of GCN, these irrelevant tokens can potentially affect the graph prompt learning process and consequently affect the overall performance. Therefore, the two modules, concept learning with token weights and graph prompt learning, are closely intertwined and mutually dependent on each other.

\subsubsection{Analysis}
\begin{table}[!ht]\caption{Comparison between \ours{} and related works on scalability. Note that, $\checkmark$(vanilla) indicates can only supports two modalities and is challenging to expand to supports more modalities.}\label{tab:table2}
\vspace{3mm}
\centering
\renewcommand\tabcolsep{5.0pt} 

    \begin{tabular}{l|ccc}
    \toprule
    Method& Multiple modalities & Unified model & Prompt   \\
    \hline
    ViT&\quad&$\checkmark$&\quad                \\
    BrainT&\quad&$\checkmark$&\quad                \\
    GraphT&$\checkmark$(vanilla)&\quad&\quad              \\
    CLIP&$\checkmark$(vanilla)&\quad&\quad         \\
    MetaT&$\checkmark$&$\checkmark$&\quad          \\
    VPT&\quad&$\checkmark$&$\checkmark$          \\
    MaPLe&$\checkmark$(vanilla)&\quad&$\checkmark$    \\
    \ours{}&$\checkmark$&$\checkmark$&$\checkmark$   \\
    
    \bottomrule
    \end{tabular}

\end{table}

\noindent \textbf{Multi-modalities}. We investigate the effectiveness of individual modalities and multi-modalities by reporting the performances in
Fig. \ref{fig:ab}. Concretely, our method considering multi-modalities (\eg MRI and PET in ADNI) obtains substantial improvements across all metrics, compared to only considering individual modalities. This suggests that the complementary information captured by different modalities can enhance the overall performance of the model.

\noindent \textbf{Interpretability}.We further investigate the interpretability of our method. To achieve this, we scale the heat maps and superimpose them on the corresponding positions of the original images. signifying higher weights and darker shades of blue signifying lower weights. 
Specifically,  the x-axis represents different positions within the same plane, while the y-axis represents different planes. 
Firstly, it is evident that our method effectively eliminates the weights associated with almost all irrelevant patches. 
Secondly, Fig. \ref{fig_vis} indicates that our method is capable of identifying the critical regions within the brain, such as hippocampal and parahippocampal. Moreover, its interpretability has been validated by clinicians.

\noindent \textbf{Scalability}. We further conduct a comparison between \ours{} and related works in terms of scalability. As shown in Table \ref{tab:table2}. we evaluate three aspects of scalability, \ie multiple modalities, unified model, and prompt. In particular, considering the inherent multimodality of medical data, we first evaluate the model's ability to handle multiple modalities.  Note that, $\checkmark$(vanilla) indicates can only support two modalities (\eg image and text) and is challenging to expand to support more modalities. Secondly, we evaluate whether the method applies a unified model for encoding multimodal medical imaging, as employing a unified model facilitates efficient inference and deployment. Thirdly, we evaluate whether the method applies prompt for fine-turning, as prompt-driven fine-turning is beneficial for enhancing the interactivity and effectiveness of the method. Comparing \ours{} with all comparison methods,  we can observe that our method achieves excellent scalability and flexibility.

\begin{figure*}[!t]
\centering
\subfigure[NC]{\includegraphics[height=1.8in, trim={2.5cm 0 1cm 2cm},clip]{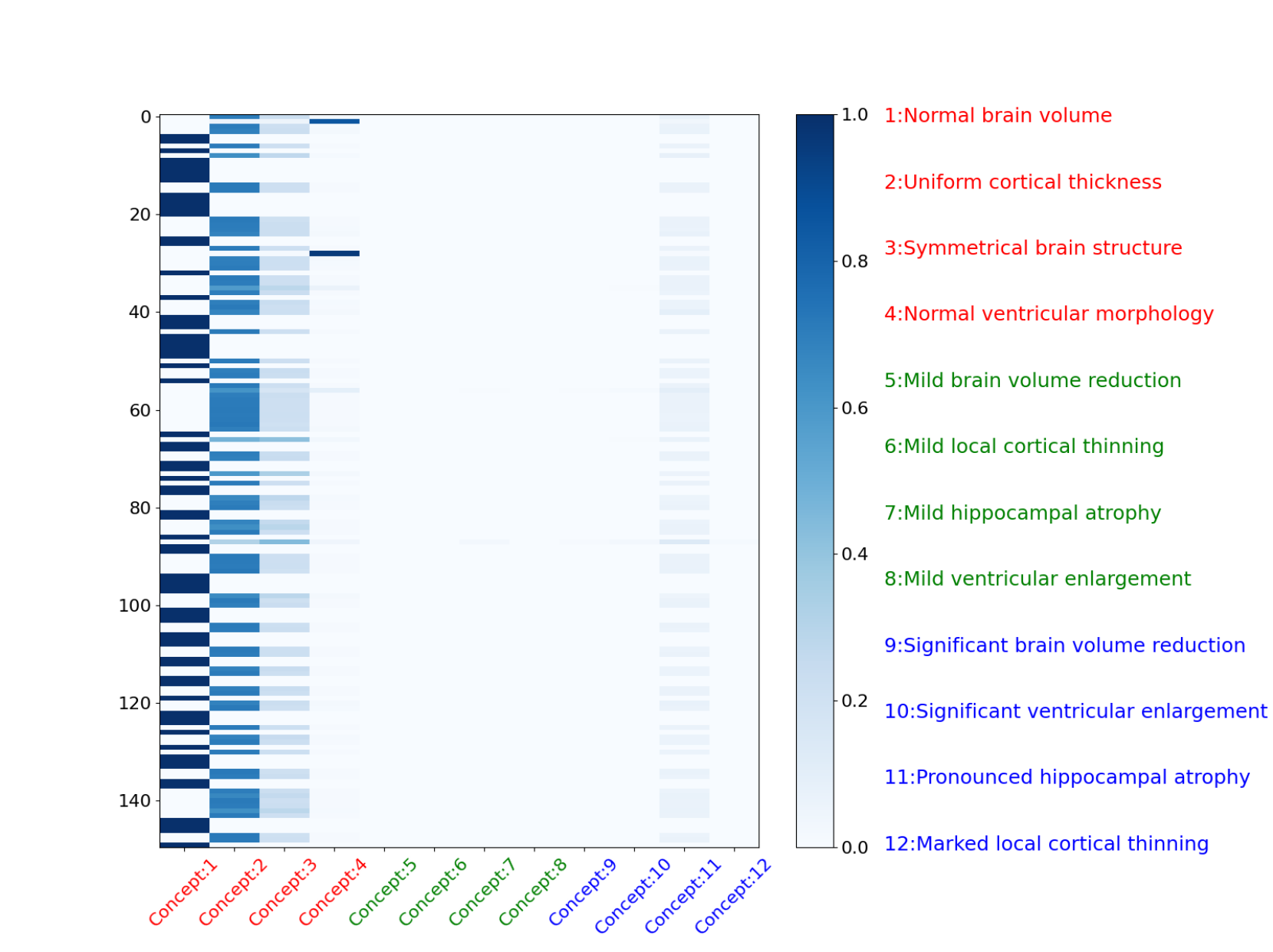}}
\subfigure[LMCI]{\includegraphics[height=1.8in, trim={2.5cm 0 1cm 2cm},clip]{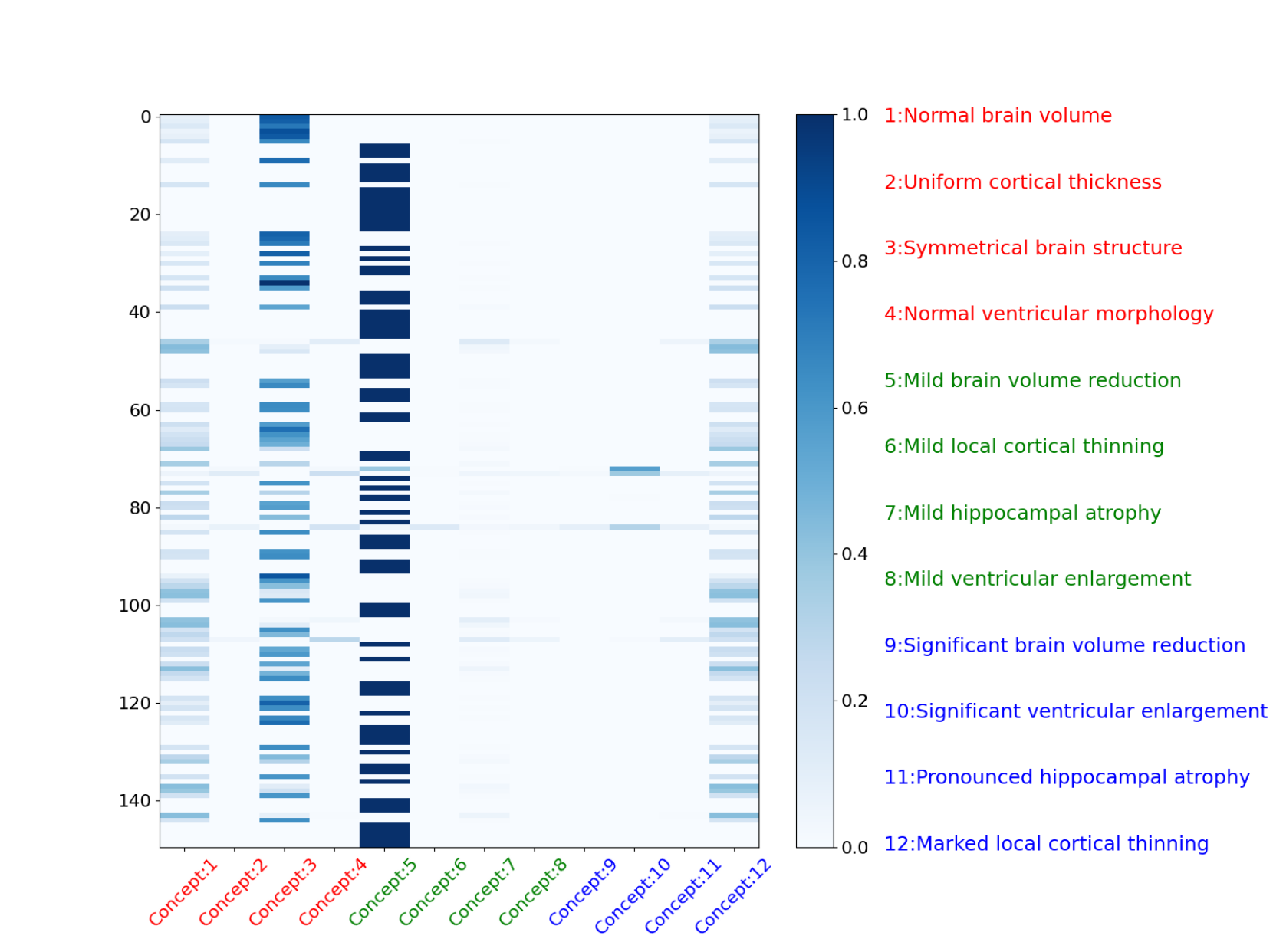}}
\subfigure[AD]{\includegraphics[height=1.8in, trim={2.5cm 0 1cm 2cm},clip]{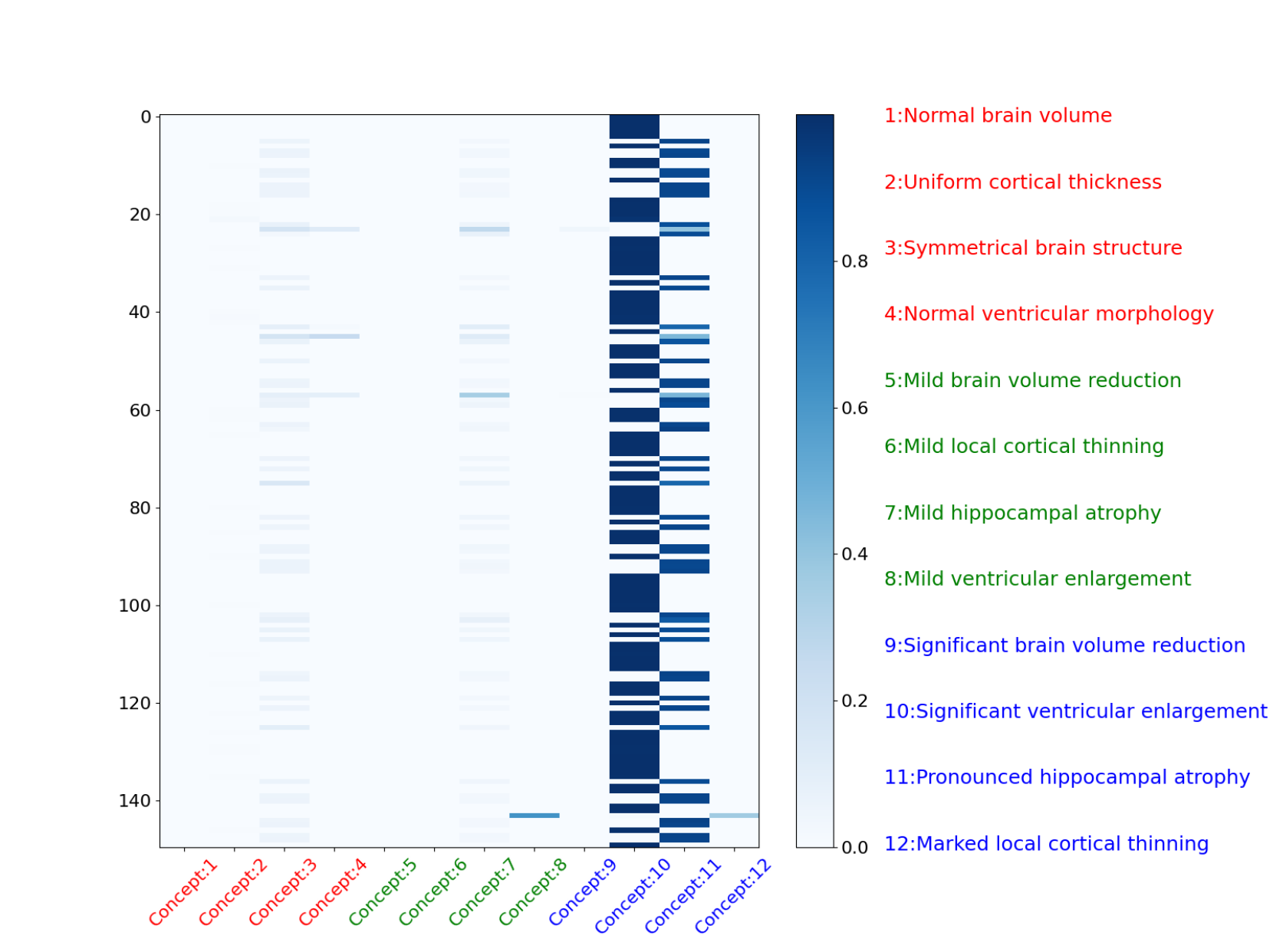}}
\caption{The visualization of concept-similarity graph on the ADNI dataset. The horizontal and vertical axes represent concepts and tokens. Different colors represent concepts belonging to different categories. The red texts represent concepts related to NC,  the green texts represent concepts related to LMCI, and the blue texts represent concepts related to AD.}
\label{fig:2-1}
\end{figure*}

\begin{figure}[!htb]
\centering
\includegraphics[scale=0.39, trim={3.2cm 3cm 3cm 4cm},clip]{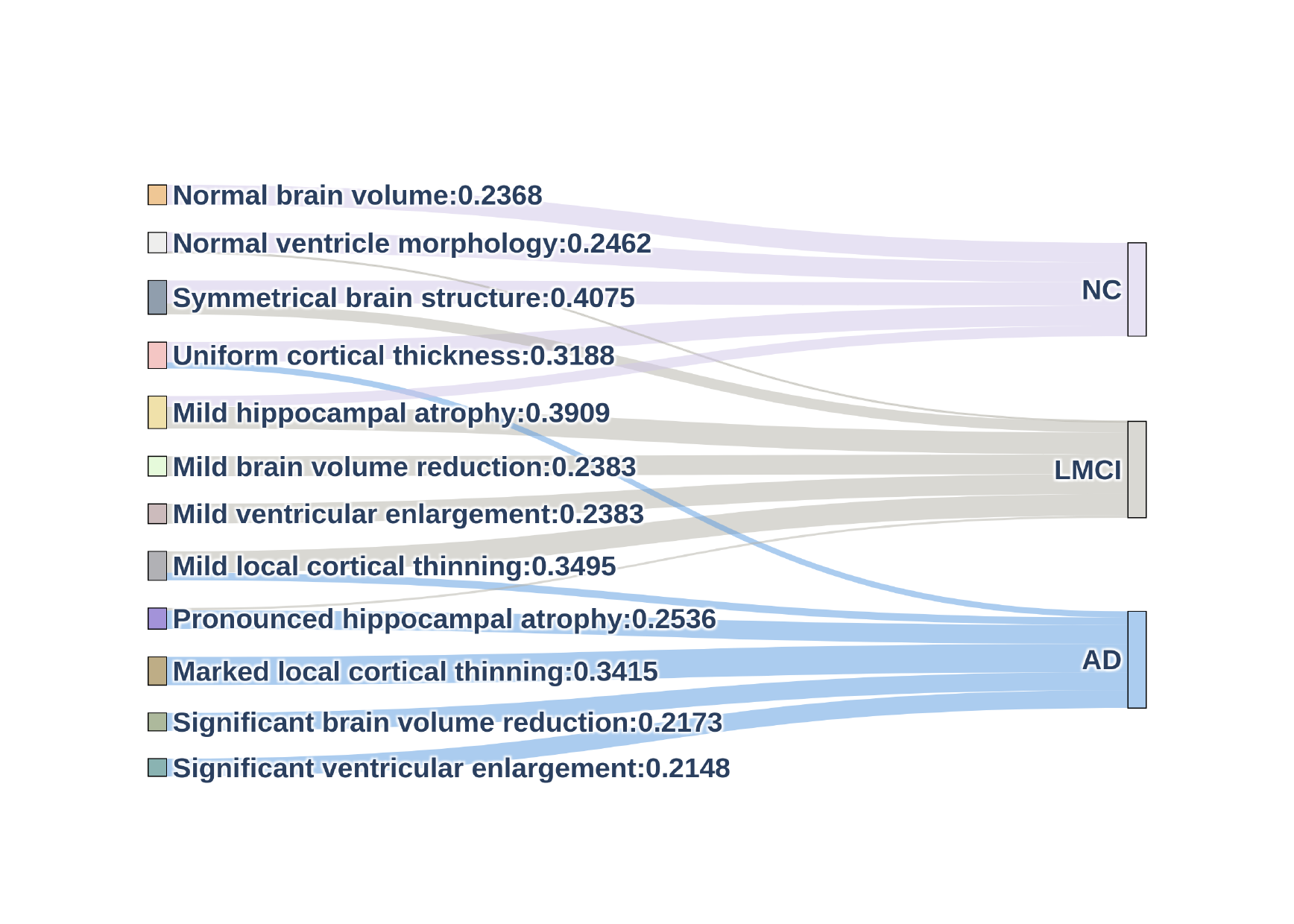}
\caption{The visualization of the quantified impact of different concepts on the ADNI dataset. The concepts are shown on the left side, while classes are shown on the right side. The width of the lines corresponds to the magnitude of the weights, and the values indicate the specific weight values.
} 
\label{fig:3-1}
\end{figure}

\noindent \textbf{Semantic similarity graph}. To improve the interpretability of \ours{}, we visualize the semantic similarity graph of samples belonging to different classes in Fig. \ref{fig:2-1}. Specifically, the semantic similarity graph is sparse. The reason could be that those concepts naturally have low similarity or limited overlap. Moreover, sparse semantic similarity graphs are easier to interpret as they highlight the most significant concepts. For example, in the middle sub-figure, the model infers the sample as LMCI category based on the fifth concept.

\noindent \textbf{Concept weights}. 
To quantify the weights of concepts, we conducted a statistical analysis on the number of times concepts were activated by samples belonging to different classes, and visualized the results using the Sankey diagram in Fig. \ref{fig:3-1}. Firstly, the weights of concepts can accurately represent their corresponding categories. In the top side of Fig. \ref{fig:3-1}, it can be observed that the top four concepts, which represent NC-related concepts, have the highest probability of being activated in NC samples. Moreover, concepts related to hippocampal atrophy tend to have higher weights and proportions in both LMCI and AD categories. This finding aligns with prior knowledge that hippocampal atrophy is associated with cognitive decline in patients with LMCI and AD.  
Secondly, There may be instances where samples from one category activate concepts belonging to other categories. For example, LMCI samples might activate concepts associated with the NC category. This could be due to individual variations, where pathological samples may not satisfy all the concepts.
It is worth noting that NC samples rarely activate concepts related to AD, while AD samples may activate concepts associated with NC. This finding aligns with clinical expectations, as NC samples typically do not exhibit any AD-related symptoms. However, some AD samples may still satisfy some concepts associated with NC. Moreover, LMCI samples can originate from concepts related to both NC and AD. This finding aligns with clinical expectations, as LMCI represents an intermediate state between NC and AD.

\section{Conclusion}
In this paper, we proposed a graph prompt learning fine-turning framework for neurological disorder diagnosis, 
by jointly considering the impact of irrelevant patches as well as the structural information among tokens in multimodal medical data. 
Specifically, we conduct concept learning, aiming to reduce the weights of irrelevant tokens according to the semantic similarity between each token and disease-related concepts.
Moreover, we conducted graph prompt learning with concept embeddings, aiming to bridge the gap between multimodal  models and neurological disease diagnosis. 
Experimental results demonstrated the effectiveness of our proposed method, compared to state-of-the-art methods on neurological disease diagnosis tasks.

\end{document}